%% file: main.tex
\documentclass[runningheads]{llncs}
\usepackage[T1]{fontenc}
\usepackage{adjustbox}
\usepackage{amsmath}
\usepackage{array}
\usepackage{booktabs}
\usepackage{xurl}
\usepackage{hyperref}
\usepackage{comment}
\usepackage{graphics, graphicx, float}
\usepackage{xcolor}
\usepackage{caption}
\usepackage{amssymb}
\usepackage{algorithm}
\usepackage{colortbl}
\usepackage{algorithmicx}
\usepackage{mathtools}
\usepackage{algpseudocode}
\usepackage{float}
\usepackage{amsfonts}
\usepackage{bbm}
\usepackage{multirow}
\usepackage{subcaption}

\begin{document}

\title{Semantic-Inductive Attribute Selection for Zero-Shot Learning}

\author{Juan Jose Herrera-Aranda\inst{1,2}\orcidID{0009-0009-3770-1269} \and
Guillermo Gomez-Trenado\inst{1,2}\orcidID{0000-0003-3366-6047} \and
Francisco Herrera\inst{1,2}\orcidID{0000-0002-7283-312X} \and
Isaac Triguero\inst{1,2,3}\orcidID{0000-0002-0150-0651}}

\authorrunning{J. J. Herrera-Aranda et al.}

\institute{Department of Computer Science and Artificial Intelligence, University of Granada  \and
Andalusian Research Institute in Data Science and Computational Intelligence (DaSCI)
\and School of Computer Science, University of Nottingham, Nottingham, UK. \email{\{juanjoha,guillermogomez\}@ugr.es, \{triguero,herrera\}@decsai.ugr.es}
}

\maketitle

\begin{abstract}
    Zero-Shot Learning is an important paradigm within General-Purpose Artificial Intelligence Systems, particularly in those that operate in open-world scenarios where systems must adapt to new tasks dynamically. Semantic spaces play a pivotal role as they bridge seen and unseen classes, but whether human-annotated or generated by a machine learning model, they often contain noisy, redundant, or irrelevant attributes that hinder performance. To address this, we introduce a partitioning scheme that simulates unseen conditions in an inductive setting (which is the most challenging), allowing attribute relevance to be assessed without access to semantic information from unseen classes. Within this framework, we study two complementary feature-selection strategies and assess their generalisation. The first adapts embedded feature selection to the particular demands of ZSL, turning model-driven rankings into meaningful semantic pruning; the second leverages evolutionary computation to directly explore the space of attribute subsets more broadly. Experiments on five benchmark datasets (AWA2, CUB, SUN, aPY, FLO) show that both methods consistently improve accuracy on unseen classes by reducing redundancy, but in complementary ways: RFS is efficient and competitive though dependent on critical hyperparameters, whereas GA is more costly yet explores the search space more broadly and avoids such dependence. These results confirm that semantic spaces are inherently redundant and highlight the proposed partitioning scheme as an effective tool to refine them under inductive conditions.

\end{abstract}

\input{00_intro.tex}

\input{01_related.tex}

\input{02_preliminaries.tex}

\input{03_methods.tex}

\input{04_experiments.tex}

\input{05_analysis.tex}

%\section{Discussion}

%\guille{Can we make a remarkable statement about the results? Otherwise this section can be omitted.}

\section{Conclusion}
\label{sec:conclusion}

In this work we addressed a central limitation of inductive ZSL: the absence of semantic information from unseen classes, which hinders the refinement of semantic spaces. We introduced a \(k\)-class stratified partitioning scheme that simulates unseen conditions using only seen data, enabling a controlled evaluation of attributes and improving the generalisation of a simple baseline such as SAE.

Within this semantic-inductive framework, we analysed two complementary strategies, RFS and GA, that apply the partitioning in different ways. Across five diverse datasets, covering human-annotated and generated semantic spaces, as well as coarse- and fine-grained domains, both approaches deliver consistent gains, with RFS offering efficiency and stability while GA provides broader exploration at higher computational cost.

Overall, our results show that inductive feature selection is a practical and effective tool to reduce redundancy and noise in semantic spaces, and that the proposed partitioning scheme provides a solid foundation for refining ZSL models under realistic inductive settings.

% Our study establishes a deployment-faithful evaluation with an outer-fold consensus threshold and delivers strong, interpretable results; building on this, we see three focused avenues for future works. Firstly, we will design nested or hybrid partitioning schemes that automate threshold selection without sacrificing the stability and inductive, class-stratified realism achieved by our current outer-fold calibration. Additionally, we will add an explanation layer that explicitly links properties—and potential biases—of visual features to the attributes retained by our selectors, deepening interpretability and enabling targeted diagnostics. (iii) We will broaden the evaluation to stronger, computationally demanding ZSL backbones to stress-test robustness and verify that our interpretability gains transfer to state-of-the-art architectures.

\section*{Acknowledgements}

% Aquí la financiación.
This work was supported by the Strategic Project IAFER-Cib (C074/23), as a result of the collaboration agreement signed between the National Institute of Cybersecurity (INCIBE) and the University of Granada. This initiative is carried out within the framework of the Recovery, Transformation and Resilience Plan funds, financed by the European Union (Next Generation).

\clearpage

\bibliographystyle{splncs04}
\bibliography{reference}

\end{document}

%% file: 00_intro.tex
\section{Introduction}
\label{sec:intro}

General-Purpose Artificial Intelligence Systems (GPAIS) aim to transcend the limitations of traditional, fixed-purpose AI by effectively performing multiple tasks, even under dynamic and unpredictable conditions where labelled data is typically scarce or unavailable \cite{TRIGUERO2024102135}. Open-world GPAIS, in particular, faces the significant challenge of flexibly generalising to novel tasks encountered at inference time, without relying on prior direct training. In addressing these demanding requirements, Evolutionary Computation (EC), a powerful tool for designing and optimising machine learning models \cite{song2019review}-- is a promising tool to enhance GPAIS capabilities in open-world settings \cite{molina2025evolutionary}.

To address such open-world challenges, Zero-Shot Learning (ZSL) emerges as a crucial paradigm.  ZSL enables models to classify examples belonging to previously unseen classes by leveraging an intermediate semantic representation, typically composed of human-defined attributes \cite{farhadi2009describing}. In contrast to related paradigms such as open set recognition --which primarily detect out-of-distribution samples-- ZSL explicitly aims at classifying these novel examples \cite{xian18ugly}.  This work specifically focuses on inductive ZSL in image classification, where no information about unseen classes is available during training, even semantic representations.

The effectiveness of ZSL very much depends on the semantic attributes. In practice, the ideal assumption that each attribute is distinctively predictive and noise-free rarely holds true \cite{jayaraman2014unreliable}. Attributes can be noisy, redundant or weakly correlated with visual differences between classes, as illustrated by common benchmarks such as the Animals with Attributes2 (AwA2) \cite{xian18ugly} and the Caltech-UCSD Birds (CUB) \cite{wah2011cub} datasets. In AwA2, ubiquitous attributes like ``has a tail'' fail to distinguish among species, whereas CUB includes subtle or subjective attributes such as specific shades of wing colour that can be challenging to reliably detect. These problematic attributes add significant noise, undermining classification accuracy and interpretability \cite{jayaraman2014unreliable}. Their identification is essential \cite{guo2018zslas} not only to reduce noise but also to enhance interpretability, fostering greater transparency and trustworthiness \cite{arrieta2020explainable}. In ZSL, examining which attributes are retained or discarded provides valuable insights into the problem domain, potentially confirming expert intuitions about relevant descriptions \cite{xu2020improving}.

%YO REDUCIRÍA UN POCO EL PARRAFO SIGUIENTE Y LO PEGARÍA AL ANTERIOR.. VEO REPETICIONES...
%Consequently, refining the semantic attribute space is an essential task \cite{guo2018zslas}. Removing redundant or non-informative attributes not only reduces noise but also enhances interpretability, enabling models to provide clearer/simpler justifications for their decisions. A refined attribute set helps align model explanations with expert human reasoning, fostering greater transparency and trustworthiness \cite{arrieta2020explainable}. Furthermore, examining which attributes are retained or discarded provides valuable insights into the problem domain, potentially challenging or confirming expert intuitions \cite{xu2020improving}. 

\begin{figure}
    \centering
    \includegraphics[width=\linewidth]{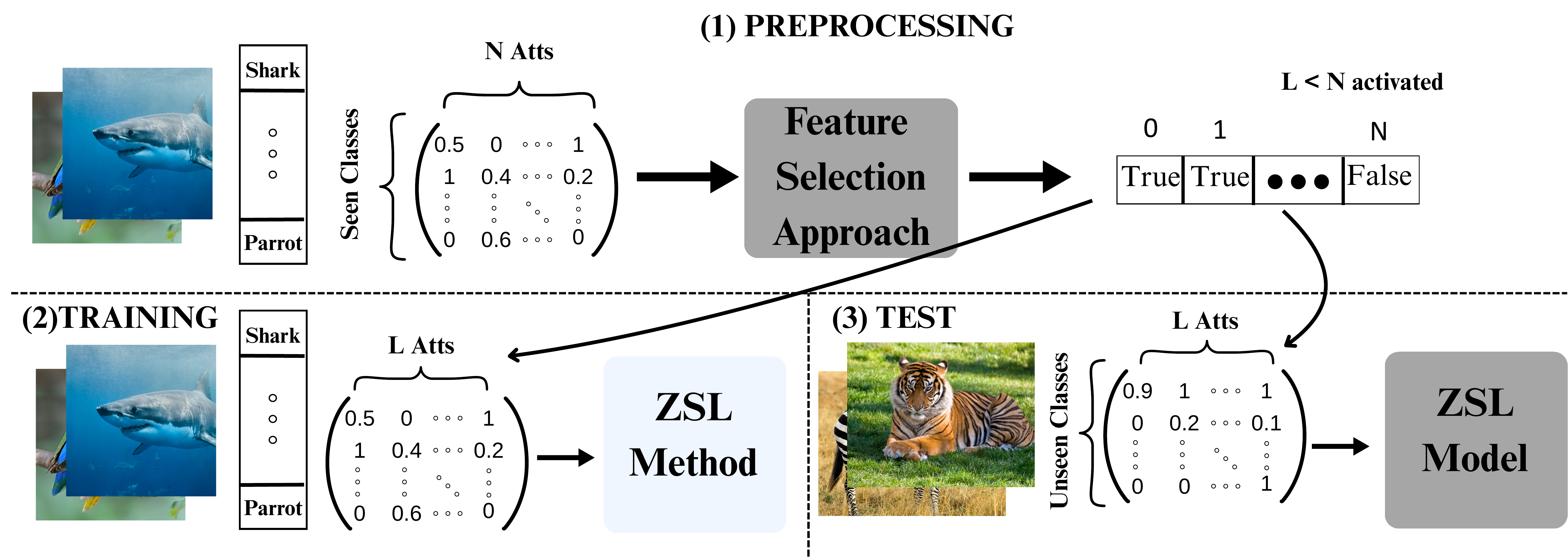}
    \caption{Workflow of an explicit semantic-inductive attribute selection. Preprocessing of the semantic space takes place without any information about the test (neither images nor their semantic information). Once completed, a ZSL method is applied as usual with the new set of chosen attributes.}
    
    %This figure summarises the preprocessing of the semantic space as a previous step to input to the model. Before carrying out the ZSL task, we make a FS preprocessing to the semantic space to remove redundant and non-informative attributes that negatively impact on the generalisation from seen to unseen classes. After selecting the attributes, they are used to train and evaluate the model. This perspective enables versatility as it can be applied to any ZSL model.}
    \label{fig:fsa}
\end{figure}

Some ZSL methods \cite{Chen23Trans,liu2024psvma+,hou2024visual} implicitly perform an attribute selection. However, each one handles attributes differently, which prevents a single strategy from being used to derive their importance. Whilst possible, extracting an attribute ranking would typically require a deep understanding of each architecture, requiring an ad-hoc procedure for each kind of model. Instead, other approaches \cite{aranda2024preliminary,herrera2025gazela} explicitly clean the semantic space prior to learning, as represented in Figure~\ref{fig:fsa}, modelling the selection of relevant attributes as a global search process. %They preprocess the semantic space prior to its use in the ZSL method. 
Thus, they might facilitate generalisation to other ZSL algorithms, providing potentially noise-free semantic spaces. The application of existing Feature Selection (FS) techniques \cite{sadeghian2025review} is not straightforward. While semantic spaces may look like tabular datasets, they are composed of one prototype per class-label. Hence, applying techniques such as embedded FS may result in overfitting and the output would be inherently biased toward seen data without guarantee of usefulness for unseen data. However, in this work, we show a strategy that adapts traditional embedded FS techniques to preprocess the semantic space, guaranteeing its generalisation to unseen classes. It is fair to mention that these techniques are not applied with the aim to reduce the number of attributes, but to find the attributes that will allow a model to generalise better to unseen classes.

%For example, \cite{Chen23Trans} uses a bidirectional attention mechanism where each attribute receives a weight per image, while PSVMA+ \cite{liu2024psvma+} learns prototypes per instance adjusted to each image.

%treating the semantic space as a step previous to the ZSL algorithm is a challenging task given that it has the structure of a tabular dataset containing a single instance per label. Thus, applying traditional Feature Selection (FS) techniques \cite{sadeghian2025review} is not a trivial task. In contrast, unlike traditional FS methods in which the purpose is reduce data dimensionality, enhance interpretability and bypass overfitting, in ZSL the goal is to improve performance on the generalisation to unseen classes rather than on the seen classes observed during training. 

Approaches that explicitly refine the semantic space \cite{jayaraman2014unreliable,xu2020improving} are usually semantic-transductive, meaning that they assume the semantic space of unseen classes is available during training, in contrast to the semantic-inductive case in which only the seen semantic space is used. The latter therefore operates in a more difficult --and perhaps realistic-- context. However, relying solely on seen classes raises a critical problem: there is no guarantee that the selected attributes will generalise well to unseen ones. To address this limitation, we propose a partitioning scheme that simulates unseen conditions with seen data, recreating a controlled ZSL scenario. This strategy allows inductive approaches to evaluate attribute relevance more reliably and enhances their potential for generalisation. 

In our previous work, we proposed two alternative approaches \cite{aranda2024preliminary,herrera2025gazela}. In \cite{aranda2024preliminary}, we employ an embedded FS algorithm to create a ranking and apply a cross-validation-based thresholding strategy to extract relevant features from the ranking. Therefore, this method relies on thresholds to find a concrete solution. To remove the need for a threshold, we proposed the use of EC to perform semantic inductive FS \cite{herrera2025gazela}. These approaches use the aforementioned partitioning scheme.

%Within the semantic-inductive case, generalisation from seen classes to unseen ones presents a difficult task. %\cite{aranda2024preliminary} t To combat it, \cite{herrera2025gazela} introduces the use of EC to conduct feature selection in the semantic space within an inductive setting to produce a unique solution. 

%Although these preliminaries works show potential results, there is still further to investigate in terms of interpretability and comparison studies of these both approaches.
% Interpretabiliaad de cada metodo
% Un estudio comparativo entre los dos
% Influencia de cada una de las partes (HAIS)

In this work, we build upon our previous approaches to (1) propose and validate a partitioning scheme that recreates a controlled inductive ZSL scenario using only seen classes; (2) deliver a thorough performance comparison across five canonical ZSL benchmarks (CUB, AwA2, SUN, aPY and FLO); (3) perform an in-depth analysis of the internal mechanisms of each method, including ablation studies; (4) analyse performance as a function of dataset typology, quantifying how granularity (fine- vs coarse-grained) and semantic provenance (human-annotated vs generated) modulate the benefits of semantic-inductive attribute selection. In our experiments, we adopt the Semantic Autoencoder (SAE) \cite{kodirov2017semantic} for its efficiency and inductive setting. Results show that both strategies achieve better results than the baseline and reach up to a fourfold improvement on aPY.

The rest of the paper is organised as follows. Section \ref{sec:related} presents related work on ZSL and preprocessing. Section \ref{sec:preliminaries} defines the notation used in this work. Section \ref{sec:methods} describes the proposed strategies to perform FS on semantic spaces for ZSL. Section \ref{sec:experiments} establishes the experimental set-up. Section \ref{sec:analysis} shows the results of different ablation studies, analysing the capabilities of the methods in depth. Finally, Section \ref{sec:conclusion} concludes the paper summarising the key contributions and lessons learned from our work.

%\guille{Falta lo de: el paper se distribuye tal que así, en la sección tal, se hace cual, en la otra lo otro, y así hasta acabar :).}

%% file: 01_related.tex
\section{Related Work}
\label{sec:related}

%\begin{figure*}[!t]    % * makes it span both columns; [!t] tries to place it at the top of a page
 % \centering
  %\includegraphics[width=\textwidth]{figs/engaging_figure.jpg}
  %\caption{\guille{Para que mi cara desaparezca de aquí, meted una figura que agite los corazones de los revisores y les devuelva la fe en la humanidad y la ciencia (no chattygpt, por favor).}}
 % \label{fig:wide}
%\end{figure*}

%\guille{Citar figura  \ref{fig:wide} en el texto.}

The design of semantic representations --describing seen and unseen classes as tuples of meaningful feature values-- has long been central to ZSL. Early seminal works employ the use of human-annotated attributes as semantic descriptors, enabling models to explain decisions transparently (e.g.,\ a zebra predicted based on attributes like stripes and four legs) \cite{farhadi2009describing,lampert2009learning}. %%%--- HE DECIDIDO QUITARLO PARA AHORRAR ESPACIO Y LO PONEMOS EN LOS EXPERIMETNOS DESPUES O CUANDO EXPLIQUE LOS DATASETS %Datasets such as Scene UNderstanding (SUN) \cite{patterson2012sun}, with 102 attributes for 717 scene categories, Animals with Attributes2 (AwA2) \cite{lampert2013attribute}, with 85 attributes for 50 animal classes and and Caltech-UCSD Birds (CUB) \cite{wah2011cub} with 312 binary attributes describing birds species, exemplify this approach.
In contrast, these manually curated attributes often suffer from incompleteness, redundancy, and difficulty in reliable visual detection \cite{jayaraman2014unreliable}. %The inherent limitations of manually defined semantic spaces have become increasingly apparent. Attributes can be misleading if ubiquitous (e.g.\ ``has tail'' among mammals) or overly subtle and subjective, particularly in fine-grained datasets such as CUB. Such issues amplify noise and reduce discriminative power, posing serious challenges for generalisation to unseen classes \cite{jayaraman2014unreliable}. 
Furthermore, these human-annotated semantic descriptions are time-consuming and difficult to elaborate.

%datasets like Caltech-UCSD Birds (CUB) \cite{wah2011cub} and Oxford Flowers (FLO) \cite{nilsback2008automated}. For example, the CUB-200-2011 dataset provides 312 binary attributes grouped into 28 semantic groups, while the Oxford 102 Flowers dataset offers discriminative yet subjective attributes for 102 flower classes. 
%By contrast, coarse‑grained datasets such as APY (Attribute Pascal and Yahoo) \cite{farhadi2009describing} amplify these drawbacks: limited attribute coverage (failing to capture fine details), noisy annotations from relying solely on bounding boxes, human subjectivity in labels, and high-dimensional visual representations. APY  comprises roughly 24,000 images across 32 classes and defines a 64‑dimensional human‑annotated semantic whose attributes are divided into three groups: space—shape (“is 3D boxy”, “is cylindrical”), parts (“has wheel”, “has wing”), and materials (“is furry”, “has wood”). Annotations collected via Amazon Mechanical Turk.

In addition to manual attributes, there is research that explores automated alternatives that require less human effort. Early approaches used text-based semantic embeddings such as Word2Vec and GloVe \cite{mikolov2013efficient,pennington2014glove}, offering scalability but sacrificing interpretability and sometimes alignment with visual cues. Yet, these mechanisms generated noisy word vectors and do not represent class semantics faithfully. Therefore, other alternatives emerged such as the employment of Large Language Models \cite{brown2020language} to automatically generate detailed class-specific descriptions \cite{shubho2023chatgpt}. Nonetheless, large language models are not noise-free when generating semantic descriptions as they usually suffer from hallucinations \cite{ye2025interpretable}.  %highlighted this issue and introduced InfZSL, which leverages large language models to automatically generate an infinite pool of phrase-level class concepts and resulting in interpretable semantic attributes, reducing hallucinations by a scoring mechanism based on entropy.

Beyond these general considerations, the semantic space, whether human-annotated or generated, introduces failure modes that have a significant impact on ZSL \cite{liu2024psvma+}. %Attributes live at mixed granularities—from low-level, directly observable cues (e.g., colour, texture) to high-level, abstract properties (e.g., behaviour)—so learning under a single granularity leads to mismatched visual–semantic links. 
The same attribute can manifest very differently across instances (e.g., a dolphin’s tail vs. a rat’s tail), creating semantic ambiguity and vague visual patterns tied to a given attribute. In combination, these factors yield insufficient visual–semantic correspondence and ambiguous predictions, limiting transfer to novel classes. This casts evidence for the importance of having a refined and representative semantic space.

Some approaches deal with the semantic space implicitly, relying on attention mechanisms or multi-granularity adaptations to down-weight irrelevant cues while amplifying informative ones \cite{chen2024progressive,Chen23Trans,liu2024psvma+,zhou2023dvgs}. These methods improve flexibility but do not output a transferable subset of attributes, sacrificing interpretability for model-specific performance. Moreover, extracting attribute importance and rank it from them to use it for others is not feasible due to several factors. First, inductive bias and poor generalisation of internal selection: the way the model implicitly treats the semantic space is linked to the specific architecture. Each method optimises the use of attributes for its own learning criteria, incorporating a certain bias. Second, each method handles the semantic space very differently from others, making it necessary the construction of a specific and complex approach for each one. For instance, PSVMA+ \cite{liu2024psvma+} generates instance-centred prototypes, whereas VADS \cite{hou2024visual} produces dynamic prototypes that may differ even within the same class. Other methods go further by merging the semantic and visual spaces into a joint embedding, making it impossible to recover attribute-level relevance. In short, the context-dependent nature of these approaches prevents isolating a consistent set of discriminative attributes.

%PSVMA+ \cite{liu2024psvma+} and VADS \cite{hou2024visual} do not assign a fixed global weight to each attribute, but the semantic importance varies depending on instances or classes. PSVMA+ generates different semantic prototypes centred on the instances, and VADS produces dynamic semantic prototypes that may differ among instances of the same class. Consequently, it is not a trivial task to obtain attributes' importance metrics. Other methods join the semantic space and image embedding space into a new embedding space, making its treatment unfeasible. In a nutshell, the dependent nature of the context for each method makes it difficult to isolate the discriminative and informative attributes.    
%. For instance, PSVMA+ \cite{liu2024psvma+} generates instance-centred prototypes, whereas VADS \cite{hou2024visual} produces dynamic prototypes that may differ even within the same class. Other methods go further by merging the semantic and visual spaces into a joint embedding, making it impossible to recover attribute-level relevance. In short, the context-dependent nature of these approaches prevents isolating a consistent set of discriminative attributes.

Other approaches treat the semantic space explicitly, aiming to retain only the most informative attributes and generalise the selection to other ZSL models. Early works demonstrated that treating all attributes uniformly was sub-optimal, highlighting the importance of addressing uncertainty and variability in attribute reliability \cite{jayaraman2014unreliable,liu2014automatic}. These early insights laid the groundwork for subsequent explicit FS methods. 
Explicit approaches can be categorised into two settings: (i) semantic-transductive, which assume access to the semantic space of unseen classes during training, and (ii) semantic-inductive, which restrict to seen classes only and therefore represent a more realistic and challenging scenario. Within the inductive setting, methods such as ZSLAS filter attributes based on discriminative power and predictability  \cite{guo2018zslas}. However, they lack explicit mechanisms to guarantee generalisation to truly unseen classes, limiting their effectiveness. By contrast, transductive approaches incorporate mechanisms that explicitly improve transferability. For instance, IAS generates pseudo-unseen class data from the seen ones and progressively refines the attribute space based on their predictive performance \cite{xu2020improving}. While such strategies achieve competitive results, they rely on semantic information from unseen classes during training, which makes them unsuitable for inductive ZSL scenarios.

%An early work \cite{jayaraman2014unreliable} introduced random forests that explicitly model attribute reliability, dynamically adjusting each attribute's influence based on its estimated uncertainty. This methodology employs the unseen semantic space, making it semantic-transductive in contrast to semantic-inductive settings, which only use the seen semantic space.%Similarly, \cite{guo2018zslas} introduced {ZSLAS}, evaluating attributes based on discriminative power and predictability and explicitly filtering out attributes with low predictive quality. This method is semantic-inductive, however, it does not incorporate any mechanism to generalise from seen classes to unseen ones.%To improve generalisation to unseen classes, iterative and pseudo-unseen class-based methods emerged. \cite{xu2020improving} introduced iterative attribute selection (IAS), which trains a generative model (e.g.\ conditional VAE) to synthesise pseudo-unseen class data from seen classes and then iteratively selects attributes based on their predictive performance on these generated samples. This strategy achieves competitive performance. However, the attribute selection is carried out employing the semantic space of unseen classes, operating in a transductive-semantic setting. 

Operating in an inductive-semantic setting presents a more complex situation to generalise from seen classes to unseen ones, given that no unseen semantic information is available. To combat this challenge, we proposed a partitioning scheme which simulates unseen classes with seen ones --named as pseudo-unseen-- recreating a controlled ZSL scenario. Building upon this idea, our first approach presents a purely semantic-inductive embedded ranking FS method incorporating internal cross-validation (RFS) \cite{aranda2024preliminary}. A feature ranking is created by embedded FS methods, and a cross-validated methodology is applied to prune attributes by evaluating subsets on folds withheld from training, reducing overfitting and ensuring that selected attributes generalise robustly to unseen classes. This proposal depends heavily on a critical hyperparameter that must be selected a priori, and there is no way to find the optimal. 
%outputs several possible solutions, depending on the level of agreement among the cross-validation folds, and selecting the best one remains an open challenge.
To overcome this limitation, we later proposed an evolutionary alternative, where attribute selection is formulated as a global search problem solved via a genetic algorithm (GA) \cite{herrera2025gazela}. Populations of attribute subsets are evolved under the same partitioning scheme, with fitness guided by pseudo-unseen accuracy. Unlike the ranking-based approach, the GA directly outputs a single refined subset without relying on critical hyperparameter choices, effectively capturing complex attribute interactions.% and outperforming deterministic selection methods, especially in high-dimensional semantic spaces.

Despite recent progress, relevant gaps remain in the literature. Most studies have focused on the transductive scenario, while the inductive case, where semantic information of unseen classes is not accessible, has received comparatively less attention. Moreover, existing works typically analyse feature selection methods in isolation, without a systematic comparison under a common framework. Finally, detailed behavioural and interpretability analyses remain scarce, even though they are key to understanding how attribute refinement contributes to generalisation. These gaps motivate our work: a thorough study of inductive feature selection in the semantic space, comparing ranking-based and evolutionary approaches across five datasets that make use of a partitioning scheme that permits operation under a semantic-inductive setting, complemented with an in-depth analysis of their behaviour and explainability.

%% file: 02_preliminaries.tex
\section{Preliminaries}
\label{sec:preliminaries}

Let $\mathcal{Y} = \mathcal{Y}^{s} \cup \mathcal{Y}^{u}$ denote the set of class labels, divided into seen classes $\mathcal{Y}^{s}$, available during training, and unseen classes $\mathcal{Y}^{u}$, observed only at test time, with $\mathcal{Y}^{s} \cap \mathcal{Y}^{u} = \emptyset$. Each class $y \in \mathcal{Y}$ is described by a unique semantic prototype vector $\mathbf{a}_y \in \mathbb{R}^{N}$ composed of $N$ semantic attributes.

The training set of seen classes is formally defined as:
\begin{equation}
    \mathcal{D}^{s} = \{(\mathbf{x}_i, \mathbf{a}_{y_i}, y_i)\}_{i=1}^{N_s},
\end{equation}
where each instance comprises visual features $\mathbf{x}_i \in \mathbb{R}^{D}$, extracted from an image, paired with the semantic prototype vector $\mathbf{a}_{y_i}$ of its corresponding class $y_i \in \mathcal{Y}^{s}$. %For each class $y \in \mathcal{Y}^{s}$, multiple visual feature vectors exist, all sharing the same semantic description $\mathbf{a}_{y}$.

The objective of our semantic refinement approach is thus to discover an attribute subset $\mathcal{A}^{*}\subseteq\{1,\dots,N\}$ that \emph{jointly} maximises $Acc_{\mathcal{Y}^{u}}$ defined in Eq.~\ref{eq:accs_u}, %, and $Acc_{\mathcal{Y}^{s}}$, 
 encouraging a semantic space that generalises to new concepts while still recognising familiar ones. In this way, prototypes are transformed from $\mathbf{a}_y \in \mathbb{R}^N$ to $\hat{\mathbf{a}}_y \in \mathbb{R}^L$ with $L < N$ through a FS projection $h(\cdot):\mathbb{R}^N \to \mathbb{R}^L$. %After training, we evaluate on two disjoint test collections:  
%(i) $\mathcal{X}^{u}=\{(\mathbf{x},y)\mid y\in\mathcal{Y}^{u}\}$, containing images of the unseen classes;  
%(ii) $\mathcal{X}^{s}=\{(\mathbf{x},y)\mid y\in\mathcal{Y}^{s},\,\mathbf{x}\notin\mathcal{D}^{s}\}$, containing \emph{unseen samples of the seen classes}.  
After finding the FS projection, we transform the semantic space, train the model and evaluate on $\mathcal{X}^{u}=\{(\mathbf{x},y)\mid y\in\mathcal{Y}^{u}\}$, containing images of the unseen classes.% as we employ ZSL setting. %In case of GZSL settings, the model must also be evaluated on $\mathcal{X}^{s}=\{(\mathbf{x},y)\mid y\in\mathcal{Y}^{s},\,\mathbf{x}\notin\mathcal{D}^{s}\}$ containing \emph{unseen samples of the seen classes}.  

%At test time, given a visual feature vector %\mathbf{x}\in\mathbb{R}^{D}$, we first predict its semantic embedding $\hat{\mathbf{a}} = g(\mathbf{x})$ through a learned visual‑semantic mapping $g:\mathbb{R}^{D}\rightarrow\mathbb{R}^{N}$. The top‑1 prediction is obtained by nearest‑neighbour search in the semantic space under cosine distance:
%\begin{equation}
%f^{u}(\mathbf{x}) = \arg\min_{y\in\mathcal{Y}^{u}} d_{\text{cos}}(\hat{\mathbf{a}}, \mathbf{a}_y),
%\end{equation}
    
%\begin{equation}
%and for the held‑out samples of seen classes:
%f^{s}(\mathbf{x}) = \arg\min_{y\in\mathcal{Y}^{s}} d_{\text{cos}}%(\hat{\mathbf{a}}, \mathbf{a}_y),
%\end{equation}
%\noindent where $d_{\text{cos}}(\mathbf{u}, \mathbf{v}) = 1 - \frac{\mathbf{u}^\top\mathbf{v}}{\|\mathbf{u}\|\|\mathbf{v}\|}$.

The corresponding accuracy is 
\begin{align}
  Acc_{\mathcal{Y}^{u}} &= \frac{1}{|\mathcal{X}^{u}|}\sum_{(\mathbf{x}, y)\in \mathcal{X}^{u}}
    \mathbbm{1}\!\big[f^{u}(\mathbf{x}) = y\big]
    \label{eq:accs_u}
    %\\[2pt]
  %Acc_{\mathcal{Y}^{s}} &= \frac{1}{|\mathcal{X}^{s}|}\sum_{(\mathbf{x}, y)\in \mathcal{X}^{s}}
  %  \mathbbm{1}\!\big[f^{s}(\mathbf{x}) = y\big]
  %  \label{eq:accs_s} 
\end{align}
with $\mathbbm{1}[P]=1$ if $P$ is true and $0$ otherwise. 

%% file: 03_methods.tex
\section{Refining the Semantic Space using Feature Selection Approaches}
\label{sec:methods}

%Motivación
As stated above, explicit identification of relevant semantic attributes may lead to a more accurate and interpretable ZSL process. However, the difficulty in a purely inductive setting lies in utilising the seen classes during the training phase to set up an environment that would allow a method to decide potentially useful features in unseen classes.

Whilst the semantic space may resemble a traditional tabular dataset, applying traditional FS methods to it is complicated as they are composed of a single prototype per class and would not have any information about the unseen classes. In the FS literature, methods are usually categorised according to the way they search for a solution (e.g., greedy vs global) or how the relevance is evaluated (e.g., filter vs wrapper vs embedded). Most supervised FS methods need multiple instances per class to estimate statistical relevance. 

In this work, we tackle inductive ZSL generalisation with a \(k\)-class stratified cross-validation partitioning (Subsection~\ref{subsec:kcv}). Using this scheme, we study two feature-selection approaches. First, a rank-based embedded method (RFS) that fits on the semantic training data to produce an attribute-importance ranking; we then refine and stabilise this ranking via a cross-validated and fold-consensus threshold that is computed across disjoint class resamplings. This refinement explicitly mitigates the brittleness of traditional single-instance-per-class ranks, where each class is represented by a single semantic prototype, by suppressing spurious, class-idiosyncratic signals and prioritising attributes that remain predictive under re-partitioning, thereby improving transfer to unseen classes (Subsection~\ref{subsec:embedded}). Second, a genetic algorithm that performs a global search over binary attribute masks, using the fold-averaged pseudo-unseen accuracy of the downstream ZSL model as the fitness signal, thereby estimating each attribute’s contribution within the full inductive pipeline (Subsection~\ref{subsec:gaa}).

% --------------------------------------------------------------------------- %
% -------------------------         CV-EFS       ---------------------------- %
% --------------------------------------------------------------------------- %
\subsection{Class Stratified Cross Validation Data Partition}\label{subsec:kcv}

\begin{figure}
    \centering
    \includegraphics[width=\linewidth]{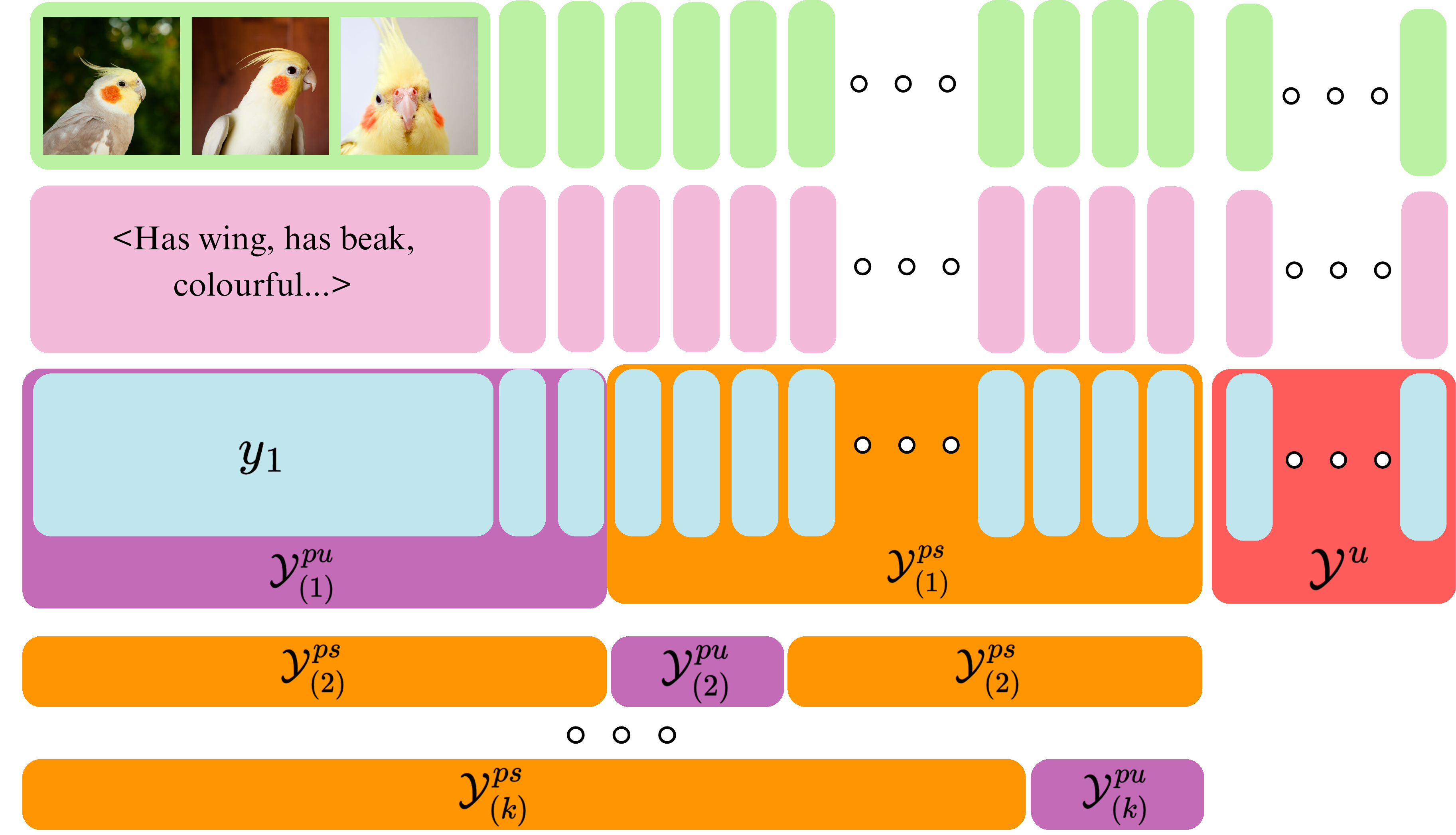}
    \caption{Workflow of the proposed class-stratified cross-validation scheme. In each fold, the set of seen classes $\mathcal{Y}^s$ is split into two disjoint subsets: pseudo-seen and pseudo-unseen. Pseudo-unseen subsets are pairwise disjoint and together cover $\mathcal{Y}^s$ combined with balance constraints. This ensures that each class is evaluated as unseen exactly once.}
    \label{fig:kcv}
\end{figure}

In semantic-inductive settings, the semantic space of unseen classes is excluded during training, which limits generalization from seen to unseen classes. To address this limitation, we introduce a class-stratified \(k\)-fold cross-validation scheme that recreates a controlled ZSL environment in each fold where unseen data is simulated. It is illustrated in Fig.~\ref{fig:kcv}. For fold $k \in \{1,\ldots,K\}$, the set of seen classes $\mathcal{Y}^s$ is partitioned into two subsets: pseudo-seen $\mathcal{Y}^{ps}_{(k)}$ and pseudo-unseen $\mathcal{Y}^{pu}_{(k)}$, such that
\begin{equation}
\label{eq:partition}
    \mathcal{Y}^{ps}_{(k)} \cup \mathcal{Y}^{pu}_{(k)} = \mathcal{Y}^{s},
    \quad
    \mathcal{Y}^{ps}_{(k)} \cap \mathcal{Y}^{pu}_{(k)} = \emptyset.
\end{equation}

Pseudo-seen classes are treated as seen data, while pseudo-unseen classes play the role of unseen data. The recreation of a controlled ZSL environment in each fold is guaranteed by the disjoint nature of $\mathcal{Y}^{ps}_{(k)}$ and $\mathcal{Y}^{pu}_{(k)}$. Furthermore, we impose three requirements to ensure a fair and unbiased evaluation:

\noindent\textbf{(Balance)} Let $n=|\mathcal{Y}^s|$, $l=\lfloor n/K \rfloor$, and $r=n \bmod K$. Each pseudo-unseen subset has size
\begin{equation}
\label{eq:balance}
    |\mathcal{Y}^{pu}_{(k)}| =
    \begin{cases}
        l+1, & \text{if } k \le r, \\[6pt]
        l,   & \text{otherwise},
    \end{cases}
\end{equation}
so fold sizes differ by at most one class. When $n$ is not divisible by $K$, the remainder $r$ is distributed by assigning one extra class to the first $r$ folds; the remaining $K-r$ folds receive exactly $l$ classes. This ensures balance evaluations with respect to the number of classes.

\noindent\textbf{(Coverage)} The pseudo-unseen subsets jointly cover the set of seen classes, ensuring that each $y \in \mathcal{Y}^s$ is treated as pseudo-unseen: 
\begin{equation}
\label{eq:coverage}
    \bigcup_{k=1}^K \mathcal{Y}^{pu}_{(k)} = \mathcal{Y}^s.
\end{equation}

\noindent\textbf{(Pairwise disjointness)} No class can appear as pseudo-unseen in more than one fold, ensuring unbiased evaluations and representation equality:
\begin{equation}
\label{eq:pairwise}
    \mathcal{Y}^{pu}_{(i)} \cap \mathcal{Y}^{pu}_{(j)} = \emptyset 
    \quad \forall\, i \ne j.
\end{equation}

All together, \eqref{eq:partition}–\eqref{eq:pairwise} define a balanced procedure that recreates a controlled ZSL setting at each fold, preventing over- and/or under-representation of any class in the unseen role.

\subsubsection{Construction of pseudo-unseen folds:}
Considering $\mathcal{Y}^s=\{y_1,\dots,y_n\}$ as an ordered list of seen classes (the order can be arbitrary). The pseudo-unseen class fold for $k=1,\dots,K$ is constructed as follows:
\begin{equation}
\label{eq:indexed}
\mathcal{Y}^{pu}_{(k)}=
\begin{cases}
\{\,y_{(k-1)(l+1)+1},\ \dots,\ y_{k(l+1)}\,\}, & \text{if } k \le r, \\[6pt]
\{\,y_{\,r(l+1)+(k-r-1)l+1},\ \dots,\ y_{\,r(l+1)+(k-r)l}\,\}, & \text{if } k > r,
\end{cases}
\end{equation}
and for each fold, we set $\mathcal{Y}^{ps}_{(k)}=\mathcal{Y}^s\setminus \mathcal{Y}^{pu}_{(k)}$. 
Therefore, by construction, Eq.~\eqref{eq:indexed} directly enforces Eqs.~\eqref{eq:balance}–\eqref{eq:pairwise}: pseudo-unseen subsets are well balanced in terms of the number of classes and differ by one class at most, and each class is treated as pseudo-unseen exactly once. 

%\begin{algorithm}[t]
%\caption{Construction of pseudo-unseen folds (implements \eqref{eq:indexed}). \guille{Qué opináis de quitarlo? me parece solo ruidoso.}
%\label{alg:kcv-construction}
%\begin{algorithmic}[1]
%\Require Ordered seen classes $\{y_1,\dots,y_n\}$ (optionally randomized), number of folds $K$
%\Ensure $\{\mathcal{Y}^{pu}_{(k)},\mathcal{Y}^{ps}_{(k)}\}_{k=1}^K$
%\State $l \gets \lfloor n/K \rfloor$; \hspace{0.5em} $r \gets n \bmod K$
%\State $c \gets 1$ \Comment cursor over the ordered list
%\For{$k \gets 1$ \textbf{to} $K$}
%    \If{$k \le r$}
%        \State $\mathcal{Y}^{pu}_{(k)} \gets \{\,y_c,\dots, y_{c+l}\,\}$ \Comment $l{+}1$ classes
%        \State $c \gets c + (l+1)$
%    \Else
%        \State $\mathcal{Y}^{pu}_{(k)} \gets \{\,y_c,\dots, y_{c+l-1}\,\}$ \Comment $l$ classes
%        \State $c \gets c + l$
%    \EndIf
%    \State $\mathcal{Y}^{ps}_{(k)} \gets \mathcal{Y}^s \setminus \mathcal{Y}^{pu}_{(k)}$
%\EndFor
%\end{algorithmic}
%\end{algorithm}

\subsection{Ranking-based Feature Selection Method}
\label{subsec:embedded}

\begin{figure}
    \centering
    \includegraphics[width=\linewidth]{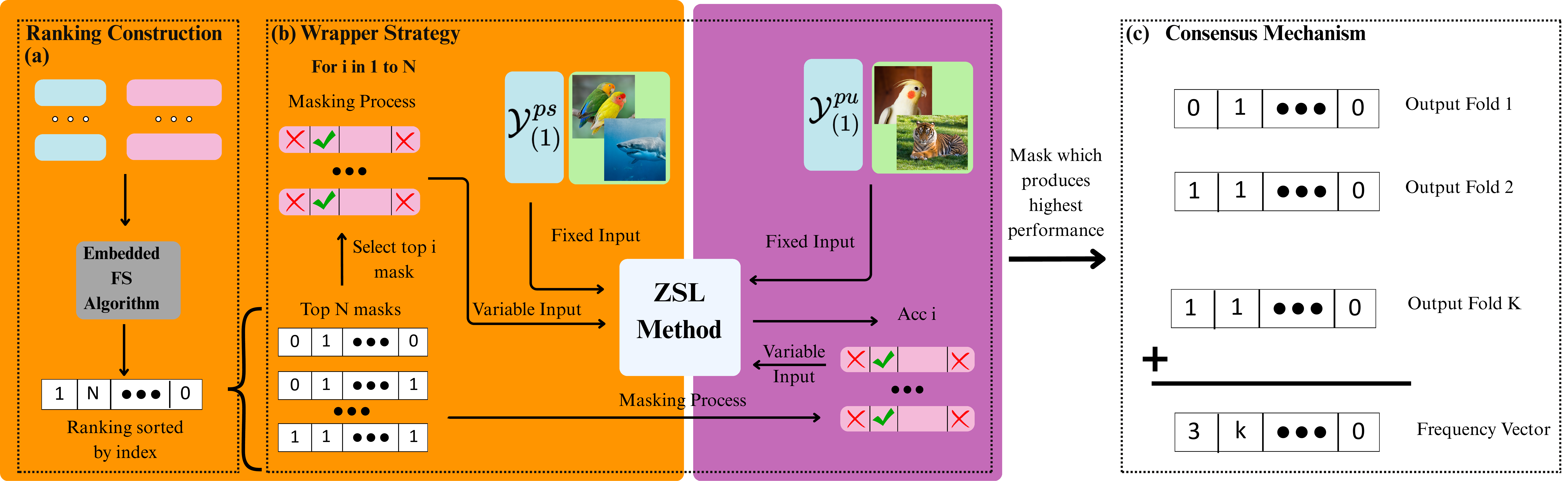}
    \caption{Workflow of the RFS methodology (We use the same colour scheme as Fig. 2). In each fold of cross-validation, an embedded feature selection algorithm produces a ranking of attributes. A wrapper strategy then evaluates successive masks (top-$i$ attributes) with the base model, and the best-performing mask is retained for that fold. After all folds are processed, a consensus mechanism combines the results into a frequency vector, from which the final subset is obtained using stability thresholds $T_1$–$T_K$ (hyperparameter established a priori).}
    \label{fig:hais}
\end{figure}

% Plantemaos un problema 
A direct way to refine the semantic space is to take advantage of the fact that, despite its limitations, it can be treated similarly to a tabular dataset and apply embedded FS techniques to it. As there is only one instance per label, the performance of these methods may be limited by overfitting. However, they still yield an attribute ranking that reflects their discriminative capacity over the seen classes. This ranking provides a first step to filter out redundant or less relevant information. Nevertheless, it remains inherently biased toward the seen data and, most importantly, offers no guarantee of usefulness for unseen classes, the very essence of ZSL.

% Explicamos RFS y como soluciona el probalme anterior
To mitigate this problem, we use the class-stratified cross-validation procedure described in Sec. \ref{subsec:kcv}. Fig.~\ref{fig:hais} illustrates this strategy. For each fold, a ranking is created (Fig.~\ref{fig:hais}(a)) and explored incrementally using a wrapper approach (Fig.~\ref{fig:hais}(b)). We start with the top-ranked attribute, train the base model on $\mathcal{Y}^{ps}_{(k)}$ and evaluate it on $\mathcal{Y}^{pu}_{(k)}$. At each iteration, a new attribute from the ranking is added and the process is repeated. The optimal subset is chosen as the one that maximises accuracy on $\mathcal{Y}^{pu}_{(k)}$. Although several strategies can be used to select an optimal subset, this wrapper approach directly optimises the true ZSL objective (performance on $\mathcal{Y}^{pu}_{(k)}$) rather than surrogate criteria on seen data. Moreover, it calibrates the embedded ranking to the base model and avoids arbitrary cut-offs, since the subset size is chosen by $Acc_{\mathcal{Y}^{pu}_{(k)}}$ via incremental walk. 

Once a solution has been obtained in each fold, it is necessary to consolidate the information in order to build a final set of attributes. To achieve this, a consensus mechanism is applied (Fig.~\ref{fig:hais}(c)): each attribute is counted according to the number of folds in which it appears in the optimal subset, and it is only retained if its frequency exceeds a threshold $T_i$, which is a hyperparameter established \textit{a priori}. This procedure acts as a regulariser, pruning attributes that look useful only under particular splits, reducing noise and redundancy in the semantic space and improving accuracy on truly unseen classes. An ablation study about the different components is conducted in Sec. \ref{subsec:hais-ablation}. 

%Limitaiocnes
The main advantage is that using embedded FS to build the initial ranking is fast, although it can overfit because there is only one prototype per class. Nonetheless, a ranking alone is not a solution: we must search for a subset, and we do it iteratively against pseudo-unseen classes to adapt the selection to ZSL. In practice, this also means (i) dependence on the initial embedded ranking, (ii) the need to pick a consensus threshold $T_i$ with no single principled choice (often yielding several plausible solutions), and (iii) multiple train/validate cycles across folds. On the positive side, the cross-validated wrapper calibrates the ranking to the base model and helps control obvious overfitting, and the final attribute sets are compact and interpretable.

% --------------------------------------------------------------------------- %
% -------------------------            GA        ---------------------------- %
% --------------------------------------------------------------------------- %
\subsection{Feature Selection using Genetic Algorithms} \label{subsec:gaa}

% Proponmos GA como alternaiva
Rather than attempting to infer attribute importance directly from the semantic space, we also evaluate it through the behaviour of a ZSL algorithm when specific attributes are included or excluded. We frame attribute selection as a global combinatorial optimisation problem and employ a GA to search for the optimal binary mask defining the selected subsets \cite{herrera2025gazela}. 

% ExplicaciÓn y comparacin con RFS para enfatizar el uso de esta alternativa
Each individual encodes a subset of attributes as a binary vector of length $N$. Each one is evaluated under the class-stratified cross-validation scheme introduced in Sec.~\ref{subsec:kcv}. In every fold we train the base ZSL model on $\mathcal{Y}^{ps}_{(k)}$ using only the selected attributes and validate on $\mathcal{Y}^{pu}_{(k)}$. The fitness is the average pseudo-unseen accuracy across folds, directly aligning the search with the ZSL objective. Unlike the ranking-based approach, this method does not rely on the election of critical hyperparameters such as the thresholds or complex mechanism such as the inter-fold consensus as in the previous methodology and returns a single subset optimised end-to-end for generalisation. 

% Le damos importancia sobre otras tecnicas clasicas
Greedy forward/backward selection or simple thresholding over a ranking often get stuck or depend on arbitrary cut-offs. However, GA explores the space globally, tolerates noisy fold-wise fitness, and does not need a ranking nor a threshold. It is also easy to modify by changing genetic operators and remains model-agnostic: any inductive ZSL backbone and validation metric can be plugged into the same fitness \cite{sadeghian2025review,song2019review}.

%Limitaciones
Despite the advantages this approach has with respect to RFS, it also presents some inconveniences: it is more computationally expensive and can be sensitive to randomness, issues often reduced by running it multiple times. These limitations will be discussed in Sec.~\ref{sec:analysis}.

%% file: 04_experiments.tex
\section{Experimental Setup}
\label{sec:experiments}

In this section, we present the details of the datasets and metrics employed in Subsection \ref{subsec:dataset_eval}, as well as the methods and hyperparameter configurations used in the experiments in Subsection \ref{subsec:hyperparameter}. The source code and scripts required to reproduce the experiments are available at \url{https://kiedie.github.io/Semantic-Inductive-Attribute-Selection-for-Zero-Shot-Learning/}.

\subsection{Datasets and evaluation protocols}\label{subsec:dataset_eval}

The ZSL literature concentrates evidence on a small set of benchmarks, often reporting partial results on only a subset. We cover the full landscape and evaluate on all five canonical datasets—Animals with Attributes (AWA2)~\cite{xian18ugly}, SUN Attributes (SUN)~\cite{patterson2012sun}, Caltech–UCSD Birds (CUB)~\cite{wah2011cub}, attribute Pascal and Yahoo (aPY)~\cite{farhadi2009describing}, and Flowers (FLO)~\cite{nilsback2008automated}—so that conclusions are not artifacts of dataset choice. This suite spans coarse- and fine-grained recognition and includes both human-annotated and generated semantic spaces, enabling a stringent, diverse validation of our methods under the same inductive protocol. Table~\ref{tab:datasets} summarises images, classes, semantic attributes (with their provenance), domain, granularity, and a brief description for each dataset.

\begin{table}[htbp]
\centering
\small
\resizebox{\textwidth}{!}{%
\begin{tabular}{
  >{\raggedright\arraybackslash}p{1.5cm}
  >{\raggedright\arraybackslash}p{1.8cm}
  >{\raggedright\arraybackslash}p{1.8cm}
  >{\raggedright\arraybackslash}p{3.8cm}
  >{\raggedright\arraybackslash}p{2.4cm}
  >{\raggedright\arraybackslash}p{5.0cm}
}
\toprule
\textbf{Datasets} & \textbf{Images} & \textbf{Classes} & \textbf{Attributes} & \shortstack[l]{\textbf{Domain}\\\textbf{(granularity)}} & \textbf{Brief description} \\
\midrule
AWA2 & 37{,}322 & 50 & 85 (human-annotated) & \shortstack[l]{Animals\\(coarse)} &
Common mammals; attributes of colour, parts, behaviour, diet, and habitat. \\
SUN  & 14{,}340 & 717 & 102 (human-annotated) & \shortstack[l]{Scenes\\(fine)} &
Indoor/outdoor scenes; materials, functions/affordances, and spatial properties. \\
CUB  & 11{,}788 & 200 & 312 (human-annotated) & \shortstack[l]{Birds\\(fine)} &
Bird species; 28 semantic groups (head, wings, pattern, colour, etc.). \\
aPY  & 15{,}339 & 32 & 64 (human-annotated) & \shortstack[l]{Objects\\(coarse)} &
20 PASCAL VOC + 12 Yahoo classes; shape, parts, and materials. \\
FLO  & 8{,}189  & 102 & 1024 (generated) & \shortstack[l]{Flowers\\(fine)} &
Oxford-102; semantic attributes obtained from textual descriptions via CNN-RNN encoding. \\
\bottomrule
\end{tabular}%
}
\vspace{0.1cm}
\caption{Summary of the datasets used in our experiments. Human-annotated attributes—originally binary—are used as normalised continuous values; FLO is the exception, where attributes are automatically extracted from textual descriptions~\cite{felix2018multi}. The \emph{Domain (granularity)} column specifies whether each dataset is fine- or coarse-grained. The table also includes a brief description of each dataset’s semantic space.
}
\label{tab:datasets}
\end{table}

For consistency with prior literature, we adopt the class splits defined by \cite{xian18ugly} for training and test sets. %, except for FLO where we use the original splits.
Visual features are extracted using a pre-trained ResNet-101 backbone, producing 2048-dimensional descriptors. Our semantic representations use continuous attribute vectors provided by the respective dataset authors. Performance is reported using %,seen and 
unseen accuracy.

In Section \ref{sec:analysis} we carry out a diversity study of the population. To measure the diversity between two individuals (binary vectors) which indicates how similar they are, we use the Hamming distance described as follows:
\begin{equation}
H_{d}(u,v) = \frac{1}{N} \sum_{i=1}^N  \vert u_i - v_i \vert 
\end{equation}
If $H_{d}(u,v) = 0$, then $u = v$ and the individuals are identical. If $H_{d}(u,v) = 1$, then $u = \neg v$ and the individuals are completely different. The diversity of the population is computed as the average of all pairwise diversity scores.

\subsection{Hyper-parameters and comparison methods}\label{subsec:hyperparameter}

The hyper-parameters employed in both methodologies are shown in Table \ref{tab:hyperparameters}.

\begin{table}[H]
    \centering
    \small
    \resizebox{\textwidth}{!}{%
    \begin{tabular}{>{\raggedright\arraybackslash}p{3cm} >{\raggedright\arraybackslash}p{10cm}}
        \toprule
        \textbf{Methods} & \textbf{Hyperparameters} \\
        \midrule
        RFS &
        \begin{tabular}[t]{@{}l@{}}
            \textbf{SAE:} \textit{Lambda} = 500000 \\
            \textbf{SVC:} \textit{Kernel} = Linear; \textit{C} = 1 \\
            %\textbf{RF:} \textit{Number of Estimators} = 100; \textit{Criterion} = Gini \\
            %\textbf{LR:} \textit{Max. Iterations} = 100; \textit{Penalty} = L2; \textit{C} = 1 \\
            \textit{k} = 5; \textit{threshold} $= T_3$ \\
        \end{tabular}
        \\
        \midrule
        GA & \textbf{SAE:} \textit{Lambda} = 500000 \\
        & \textit{pop\_size $ = 50$}; \textit{generations} $= 150$; \textit{tournament\_size} $ = 3$; {\textit{crossover\_probability} $= 0.2$};\textit{mutation\_probability} $= 0.8$ \\
        \bottomrule
    \end{tabular}
    }
    \vspace{0.1cm}
    \caption{Hyperparameter Specifications}
    \label{tab:hyperparameters}
\end{table}

Both RFS and GA require a base ZSL model to evaluate candidate attribute subsets. In our case, we adopt SAE \cite{kodirov2017semantic}, chosen for two main reasons: (i) it is purely inductive, fitting the semantic–inductive setting of our study, and (ii) it is computationally efficient, which is beneficial given the large number of training and evaluation cycles required. Alongside comparing RFS and GA, we also contrast their performance with SAE, in order to assess the added value of attribute selection over using the full semantic space. SAE operates as follows:

%%%%%%% ----- SAE ----- 
During training, it learns a linear projection \(W\) with tied weights (encoder \(g(\mathbf{x}) = W\,\mathbf{x}\) and decoder \(\hat{\mathbf{x}} = W^\top \mathbf{a}\)) using pairs \((\mathbf{x}_i,\mathbf{a}_i)\). With data matrices \(X=[\mathbf{x}_1,\dots,\mathbf{x}_N]\in\mathbb{R}^{D\times N}\) and \(S=[\mathbf{a}_1,\dots,\mathbf{a}_N]\in\mathbb{R}^{K\times N}\), it optimizes a trade-off between reconstructing visual features from the semantic space and enforcing the latent code to match the semantic attributes:
$$\min_{W}\ \|X - W^\top S\|_F^2\;+\;\lambda\,\|W X - S\|_F^2,$$
where \(\lambda>0\) balances decoder reconstruction and encoder semantic fit. Setting the gradient to zero yields a Sylvester equation for \(W\):
$$S S^\top\, W \;+\; \lambda\, W\, X X^\top \;=\; (1+\lambda)\, S X^\top,$$
which can be written as \(A W + W B = C\) with \(A = S S^\top\), \(B = \lambda X X^\top\), ${C = (1+\lambda) S X^\top}$, and solved with standard methods.% The resulting \(W\) defines the visual-semantic mapping function $g:\mathbb{R}^{D}\rightarrow\mathbb{R}^{N}, \; g(\mathbf{x})=W\,\mathbf{x}$ used at test time.

Test time operates as follows: given a visual feature vector $\mathbf{x}\in\mathbb{R}^{D}$, we first predict its semantic embedding $\hat{\mathbf{a}} = g(\mathbf{x})$. The top‑1 prediction is obtained by nearest‑neighbour search in the semantic space under cosine distance:
\begin{equation}
f^{u}(\mathbf{x}) = \arg\min_{y\in\mathcal{Y}^{u}} d_{\text{cos}}(\hat{\mathbf{a}}, \mathbf{a}_y),
\end{equation}
    
\noindent where $d_{\text{cos}}(\mathbf{u}, \mathbf{v}) = 1 - \frac{\mathbf{u}^\top\mathbf{v}}{\|\mathbf{u}\|\|\mathbf{v}\|}$.

For the main comparisons, we adopt RFS with SVC and the consensus threshold \(T_3\) as the primary configuration, as it offers a strong accuracy–stability trade-off across datasets. A comprehensive ablation covering all embedded rankers and thresholds is provided in Subsection~\ref{subsec:hais-ablation}.

Concerning GA, to account for its stochastic nature, we run $20$ independent executions and average the results on the unseen data to compare with other methods. To implement it we used the \textit{DEAP} Python library, particularly, we used the \textit{eaSimple} algorithm \cite{fortin2012deap}. All experiments were executed on a server with the following characteristics: 19 nodes, each with two 2.2~GHz (12-core) Intel Xeon Silver 4214 processors with Intel Turbo Boost Technology 2 and Hyper-Threading; 11~MB of cache memory per processor; 256~GB DDR4 2666 ECC RAM; 1~NVMe SSD 512~GB Enterprise at 3\,400~MB/s for writing, and 320~TB resilience; running Ubuntu~20.04 LTS.

%% file: 05_analysis.tex
\section{Results and analysis}
\label{sec:analysis}

In this section, we provide a comparison of the performance among the methods (Subsection \ref{subsec:comparison-baselines}). Then, we conduct an in-depth study of the RFS behaviour, examining its different components, the various embedded FS methods, and the thresholds (Subsection \ref{subsec:hais-ablation}). Finally, we exhaustively analyse the behaviour of the GA approach in terms of attribute attribute frequency convergence, fitness convergence, and diversity (Subsection \ref{subsec:ga-ablation}).

\subsection{Comparison with Baseline Methods}
\label{subsec:comparison-baselines}
To assess the effect of preprocessing the semantic feature space, we compare RFS and GA against the baseline SAE. As summarised in Table~\ref{tab:comparison}, attribute selection typically improves unseen accuracy across datasets. The notable exception is SUN, where the margins between SAE and RFS are narrow. SUN spans a wide variety of classes (717) but offers only a limited set of fine-grained, human-annotated attributes (102), yielding a relatively coarse semantic vocabulary compared with its class diversity. In this regime, redundancy is scarce: RFS’s conservative, consensus thresholding therefore retains most attributes and brings little headroom for further gains, while GA’s global search prunes a small subset and achieves a greater edge. Overall, the differences on SUN are modest, consistent with the dataset’s limited scope for semantic compression.

\input{tables/exp_comp}

Additionally, the aPY dataset is known to perform poorly with SAE, as previously reported in \cite{xu2020improving}. In our case, the baseline accuracy is indeed very low. However, by applying attribute selection we achieve a dramatic improvement: RFS  accuracy represents almost a $400\%$ increase. In contrast, GA is not able to improve over the baseline. We hypothesise that RFS improves aPY because its step-by-step process first ranks the attributes and then refines them. This helps remove noisy and weakly annotated attributes that limit SAE in this small and diverse dataset. The fold-consensus threshold highlights attributes that are clearly linked to visual features, which makes the semantic space cleaner and explains the large improvement we observe. By contrast, GA depends on a strong, stable fitness signal; with a very low baseline and high variance on aPY, its global search becomes noisy and prone to over-pruning, so it fails to beat the baseline.

Overall, the results reveal clear redundancy in semantic spaces, even when they are human-annotated, and both RFS and GA exploit this redundancy in complementary ways. RFS tends to preserve larger subsets of attributes, favouring stability through its incremental ranking procedure. GA, by contrast, explores combinations more freely and prunes more aggressively, exposing overlapping or weakly discriminative descriptors. This explains why GA solutions are generally smaller, while RFS subsets are larger but still competitive.

The extent of reduction, however, varies greatly across datasets. In human-annotated spaces the pruning required is usually moderate, though not uniform: AWA2 benefits from only a small reduction, while CUB reaches its best results with a more substantial cut. These patterns do not align neatly with the coarse/fine taxonomy, suggesting that annotation quality and coverage play a stronger role than granularity alone. By contrast, generated spaces such as FLO demand extreme pruning (around $-65\%$) to achieve competitive performance, which reflects their noisier and more redundant nature.

When comparing GA and RFS, neither method can be considered strictly superior. As we will see later, GA is computationally more demanding but has the advantage of exploring the search space more broadly and does not rely on fixed hyperparameters such as stability thresholds. In contrast, RFS is far more efficient and, as the results show, remains highly competitive with GA. Its main limitation is the dependence on several critical hyperparameters, which can strongly affect performance (see Subsec.~\ref{subsec:hais-ablation}). In practice, the two approaches are complementary: GA offers flexibility and robustness at higher cost, while RFS provides a lighter alternative that can achieve comparable accuracy when its hyperparameters are well tuned.

\subsubsection{Runtime Analysis}

\renewcommand{\arraystretch}{1.2}
\setlength{\tabcolsep}{2pt}
\begin{table}[H]
  \centering
    \begin{tabular}{|l|c|c|c|c|c|c|c|c|c|c|}
    \hline
     & \multicolumn{2}{c|}{\textbf{AWA2}} & \multicolumn{2}{c|}{\textbf{CUB}} & \multicolumn{2}{c|}{\textbf{SUN}} & \multicolumn{2}{c|}{\textbf{aPY}} & \multicolumn{2}{c|}{\textbf{FLO}} \\ 
     \hline 
     & Time & \#SAE & Time & \#SAE & Time & \#SAE & Time & \#SAE & Time & \#SAE \\
    \hline
    \textbf{SAE}        & 6 & --   & 10 & --   & 7 & --   & 7 & --   & 31 & -- \\
    
    \textbf{GA}         & 19581 & 7113  & 79175 & 7105 & 30838 & 6986 & 14355 & 7093 & 206223 & 7138 \\
    \textbf{RFS-SVC}    & 2259 & 300  & 20842 & 287 & 2832 & 385 & 1306 & 40 & 79009 & 4995 \\
    \hline
    \end{tabular}
  \vspace{0.1cm}
  \caption{Execution time in seconds (Time) and number of SAE trainings (\#SAE) across datasets. SAE trainings are the main computational bottleneck. For RFS, this number scales with the dimensionality of the semantic space: fewer attributes (aPY) require fewer trainings, while larger spaces (FLO) demand substantially more. For GA, the stopping criterion is the number of generations; however, we implemented a cache to avoid retraining identical subsets, which explains the slight variations across datasets.}
  \label{tab:runtime_unified}
\end{table}
\renewcommand{\arraystretch}{1}

Table~\ref{tab:runtime_unified} compares the execution time and the number of SAE trainings, which is the main computational bottleneck. It can be appreciated that the genetic approach is several orders of magnitude slower. This is expected, since GA explores a much larger portion of the search space. %Nevertheless, GA has the advantage of not requiring the manual specification of a critical hyperparameter such as the threshold in RFS, offering a more flexible and bigger exploration of the solution space at the cost of much more higher runtime.
Furthermore, it is worth highlighting the importance of employing a highly efficient base model such as SAE within these approaches, since adopting more computationally expensive ZSL methods would render the execution of both RFS and genetic strategies practically infeasible.

\subsection{Analysis in depth of RFS}
\label{subsec:hais-ablation}

The RFS methodology has two phases: (i) rank attributes by importance using embedded FS methods, and (ii) select an optimal subset through cross-validation with stability thresholds ($T_5$–$T_1$). In this ablation, we evaluate the role of the embedded ranking and of the thresholds. Then, we briefly analyse the impact of dataset granularity and semantic space type. Finally, we make an overall conclusions. Results are depicted in Table~\ref{tab:ablation_hais} showing the number of selected attributes and the unseen accuracy per dataset, method, and threshold.

%htbp

\begin{table}[H]
  \centering
  \resizebox{0.9\linewidth}{!}{%
  \begin{tabular}{l | l | rr | rr | rr | rr | rr }
    \toprule
    \multirow{2}{*}{DATASETS} & \multirow{2}{*}{FS METHOD} 
      & \multicolumn{2}{c|}{$T_5$} 
      & \multicolumn{2}{c|}{$T_4$} 
      & \multicolumn{2}{c|}{$T_3$} 
      & \multicolumn{2}{c|}{$T_2$} 
      & \multicolumn{2}{c}{$T_1$}  \\
    & 
      & Att  & $Acc_{\mathcal{Y}^{u}}$      
      & Att  & $Acc_{\mathcal{Y}^{u}}$
      & Att  & $Acc_{\mathcal{Y}^{u}}$
      & Att  & $Acc_{\mathcal{Y}^{u}}$
      & Att  & $Acc_{\mathcal{Y}^{u}}$    \\
    \midrule
    \multirow{4}{*}{AWA2}& RF     &  28 & 35.12 &  64 & 42.64 &  75 & 42.53 &  77 & 42.59 &  81 & 40.53  \\
    & SVC    &  65 & 36.59 &  78 & \textit{42.16} &  82 & \textbf{43.98} &  82 & \textit{43.98} &  84 & \textit{42.07}   \\
    & LR     &  12 & 31.85 &  45 & 39.76 &  70 & 39.72 &  85 & 40.36 &  85 & 40.36  \\
    & Random &  26 & \textit{37.74} &  67 & 41.57 &  70 & 41.27 &  72 & 41.11  &  85 & 40.36  \\
    \addlinespace
    \hline
    \addlinespace
    \multirow{4}{*}{SUN}& RF     &  98 & 46.81 & 100 & 46.81 & 100 & 46.81 & 101 & \textit{47.01} & 102 & 47.01  \\
    & SVC    &  92 & 46.11 &  98 & 46.60 & 100 & 46.88 & 101 & 46.94 & 102 & 47.01  \\
    & LR     &  57 & 40.90 &  95 & \textit{47.15} & 101 & \textbf{47.71} & 102 & 47.01 & 102 & 47.01  \\
    & Random & 102 & \textit{47.01} & 102 & 47.01 & 102 & 47.01 & 102 & 47.01 & 102 & 47.01  \\
    \addlinespace
    \hline
    \addlinespace
     \multirow{4}{*}{CUB}& RF     & 188 & 38.46 & 293 & 39.50 & 308 & 39.03 & 312 & 38.79 & 312 & 38.79  \\
    & SVC    & 256 & \textit{38.90} & 265 & 39.30 & 287 & \textit{39.16} & 301 & \textit{39.00} & 311 & \textit{38.86}  \\
    & LR     & 171 & 38.05 & 271 & \textbf{39.81} & 309 & 38.96 & 312 & 38.79 & 312 & 38.79  \\
    & Random & 142 & 36.84 & 190 & 37.18 & 190 & 37.18 & 218 & 37.21 & 265 & 38.73  \\
    \addlinespace
    \hline
    \addlinespace
    \multirow{4}{*}{aPY}& RF     &  31 & \textbf{21.87} &  44 & \textit{21.39} &  47 & \textit{21.64} &  51 & \textit{21.38} &  58 & \textit{21.42}  \\
    & SVC    &  11 & 14.22 &  32 & 15.50 &  43 & 21.27 &  53 & 17.55 &  55 & 18.55 \\
    & LR     &   4 &  5.50 &  25 &  5.50 &  52 &  5.50 &  59 &  5.50 &  64 &  5.50  \\
    & Random &  20 &  5.50 &  20 &  5.50 &  62 &  5.50 &  64 &  5.50 &  64 &  5.50  \\
    \addlinespace
    \hline
    \addlinespace
    \multirow{4}{*}{FLO}& RF     & 571 & 36.71 & 873 & 36.97 & 991 & 37.58 & 1020 & 37.75 & 1024 & 37.66  \\
    & SVC    & 178 & \textit{42.60} & 282 & \textit{42.51} & 364 & \textbf{42.60} &  534 & \textit{41.13} &  832 & \textit{40.17} \\
    & LR     &  83 & 34.55 & 411 & 36.80 & 780 & 38.10 &  988 & 38.18 & 1023 & 37.84  \\
    & Random & 345 & 37.49 & 525 & 37.58 & 635 & 36.97 &  995 & 38.18 & 1005 & 38.01  \\
    \bottomrule
  \end{tabular}%
  }
  \caption{ Accuracy ($Acc_{\mathcal{Y}^{u}}$) and number of selected attributes (Att) for each stability threshold ($T_5$–$T_1$) and FS method (RF, SVC, LR, Random). Best results in each dataset are highlighted in \textbf{bold}. Best results at each threshold for each dataset are pointed out in \textit{italic}, except when all attributes are selected. In the case of a tie, the one with fewer attributes prevails.}
  \label{tab:ablation_hais}
\end{table}

\subsubsection{Embedded Rankings Analysis:}To evaluate the role of the ranking selection, we compare three embedded FS methods (Random Forest , RF; Liner Regression, LR; and Support Vector Machine, SVC) with a random ranking. Specifically, in the random-based setting, we replace the embedded method with a uniform random ordering of all attributes and then run the same subset selection as usual. We keep the cross-validation protocol and the subset-selection procedure fixed, and change only the ranking step. We do not remove or alter cross-validation because the optimal subset search depends on it, and changing it would change the nature of the methodology.

\begin{figure}
    \centering
    \includegraphics[width=\linewidth]{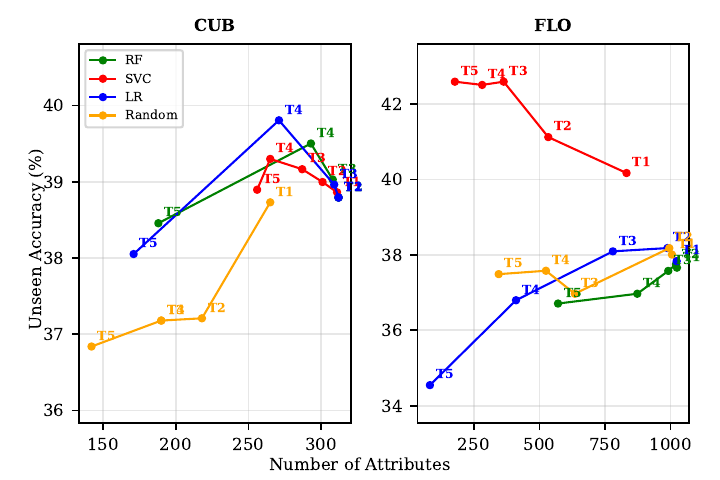}
    \caption{Relation between the number of selected attributes and unseen accuracy for different thresholds ($T_1$–$T_5$). Results are shown for embedded methods (RF, SVC, LR) compared to a random ranking on representative datasets. The plots illustrate how stricter thresholds reduce the attribute set while affecting accuracy differently depending on the ranking strategy.}
    \label{fig:attvsT}
\end{figure}

When comparing the three embedded methods with random ranking, we observe that embedded methods consistently dominate the random approach across all datasets. Not only does the best performance of each method surpass the random solution, but also in the majority of cases, embedded solutions exceed the random approach. For instance, we can appreciate in Fig.~\ref{fig:attvsT} that in FLO, all random solutions always underperform compared with other methods, where SVC clearly outshines. On the other hand, while random can beat particular embedded/threshold combinations (e.g., RF on CUB), it never surpasses the majority of the other embedded choices.

The magnitude of the improvement varies depending on the dataset. Gains are subtle when many attributes are broadly informative (i.e., most attributes are chosen), such as SUN, and larger when not all the attributes are as informative and embedded rankings significantly differ from random ranking in terms of selected attributes (e.g., FLO) . In the most balanced thresholds (discussed later), Fig.~\ref{fig:Tvsper} shows how the embedded FS methods outperform the random approach. The trend is consistent: rankings based on built-in characteristics offer superior and more stable performance than random ordering. 

These results also validate the class-stratified \(k\)-fold protocol itself: it lets embedded selectors—originally not designed for the single-prototype-per-class regime—express their genuine signal while curbing overfitting and other pathologies. In practice, the partitioning converts raw embedded rankings into robust inductive ZSL subsets, aligning selection with the unseen-class objective rather than quirks of any single split.

\subsubsection{Thresholds Analysis:} To evaluate the impact of the thresholds we simply make a comparison among them when using embedded FS approaches. We treat thresholds explicitly as a hyperparameter, controlling the level of inter-fold agreement. Threshold $T_i$ contains the attributes selected in at least $i$ times across folds. A stricter consensus such as $T_5$ yields fewer attributes, while a very permissive setting such as $T_1$ often behaves close to the baseline that keeps all attributes. In some datasets such as SUN, AWA2 and CUB, $T_2$ and $T_1$ are very similar to this baseline because it retains almost every attribute. 

\begin{figure}
    \centering
    \includegraphics[width=0.8\linewidth]{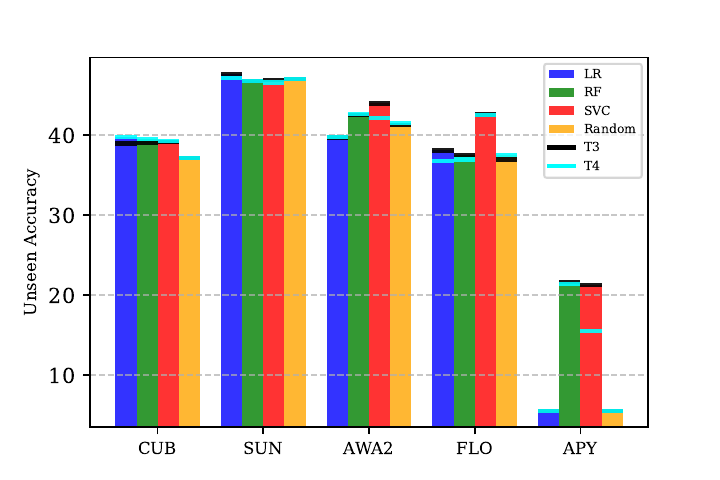}
    \caption{Comparison of unseen accuracy across datasets for intermediate thresholds $T_3$ and $T_4$. Results are reported for the three embedded methods and a random approach, showing how different thresholds influence the balance between attribute reduction and performance.}

    \label{fig:Tvsper}
\end{figure}

In general, the intermediate thresholds $T_3$ and $T_4$ provide the most balanced trade-off between the number of selected attributes and the achieved performance. The comparison between them is particularly illustrative: when $T_4$ is better than $T_3$, the difference is usually small, but when $T_3$ surpasses $T_4$, the gain is often more noticeable, especially with SVC (see Fig.~\ref{fig:Tvsper}). 

This behaviour highlights the challenge of cleaning the semantic space in ZSL. If the threshold is too strict, we risk discarding features that unseen classes may need. If it is too loose, we keep noisy attributes that hide the useful signal. Exploring the full range from $T_5$ to $T_1$ therefore helps to visualise this trade-off and to choose the level that generalises best. %\textcolor{green}{La siguiente frase no se si ponerla porque corresponde aqui porque estamos enfatizando el particionado: The key challenge is to ensure generalisation to truly unseen classes, since attributes that degrade performance on seen classes may prove useful for unseen ones and it is done with the partitioning scheme in \ref{subsec:kcv}.} %That is why T1 can sometimes score higher than the baseline. 

\subsubsection{Dataset Characteristics} Another aspect to consider is how dataset properties influence attribute selection. The clearest contrast appears between human-annotated and generated semantic spaces. While human-annotated datasets (AWA2, SUN, aPY, CUB) tend to be more stable and less noisy, FLO, with its very high dimensionality, requires an aggressive pruning of attributes and shows less consistency across methods (Fig.~\ref{fig:attvsT}), reflecting the redundancy of generated spaces. 

Within the human-annotated group, the picture is more nuanced. AWA2 (coarse-grained) keeps most attributes even under stricter thresholds, and SUN, despite being fine-grained—also retains the majority, likely because its 102 attributes form a compact, broadly informative set relative to its 717 classes. By contrast, aPY (coarse-grained) undergoes stronger pruning, pointing to dataset-specific redundancy or limited annotation coverage. CUB (fine-grained) shows a substantial reduction ($\approx35\%$), indicating overlap among its many detailed descriptors. Overall, these patterns suggest that granularity alone does not predict how much pruning is beneficial; the quality and coverage of the attribute annotations, and the attribute-to-class ratio, are more informative.

\subsubsection{Overall conclusions} When applying the thresholds derived from cross-validation ($T_1$–$T_5$), embedded methods provide a clear mechanism to stabilise attribute selection and in most cases lead to improved generalisation compared to using all attributes. The gains are particularly evident in datasets with higher redundancy, while in cases such as SUN, where almost all attributes are informative, the baseline is already strong and the benefits of pruning are naturally limited. Among the different settings, intermediate thresholds ($T_3$–$T_4$) generally offer the best compromise between accuracy and subset size. In our experiments, $T_3$ was adopted as the most balanced choice, and within this configuration SVC consistently delivered competitive results across datasets.

\subsection{Analysis in depth of GA Approach}\label{subsec:ga-ablation}

\begin{figure}[H]
    \centering
    \includegraphics[width=0.8\linewidth]{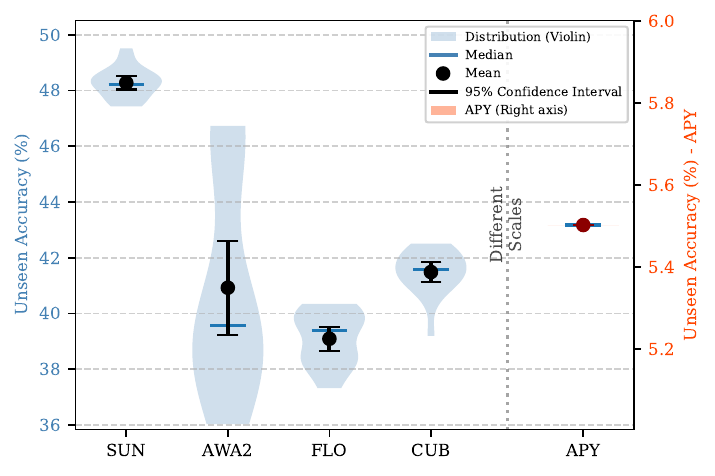}
    \caption{Violin plots of the $Acc_{\mathcal{Y}^u}$ obtained across 20 independent GA runs for each dataset. The distribution (violins) reveals the variability across executions, while dots and vertical bars denote the mean and its 95\% confidence interval. SUN and CUB exhibit stable behaviour with narrow confidence intervals, FLO shows moderate dispersion, and AWA2 displays the largest variability. In contrast, aPY shows almost no variability, with all runs converging to the same poor performance, which indicates that the GA is unable to escape the limitations of this dataset. }

    \label{fig:violin}
\end{figure}

In this section we focus on the behaviour of the Genetic Approach. The analysis is divided into two parts: (i) attribute frequency analysis and (ii) fitness and diversity analysis. From the five datasets considered in this work, we select FLO as a generated semantic space, AWA2 as a coarse-grained human-annotated dataset, and CUB as a fine-grained human-annotated dataset to carry out this analysis. As in the comparative experiments (Subsection \ref{subsec:comparison-baselines}), the study is based on $20$ independent runs of the genetic search to strengthen the robustness of the observations as it can be observed in Figure~\ref{fig:violin}, some datasets present high variability across execution; hence the need for computing the average.

\subsubsection{Attribute Frequency Analysis}

Fig.~\ref{fig:genetic-behaviour} (a) shows the selection frequency of each attribute across runs. Three main groups can be distinguished. First, consistently selected attributes (frequency $>$15), appear in almost every run; they represent the most discriminative features. Second, occasionally selected attributes (frequency 6–15,) are chosen in some executions but not in others, which suggests that their contribution is context-dependent and may rely on specific data subsets or initialisation. Finally, rarely or never selected attributes (frequency $<$6), have very low counts, indicating redundancy or little discriminative power. 

Although the mean selection frequency is similar across datasets, the distributions reveal very different behaviours. FLO exhibits a shape close to a normal distribution centred around the mean. This means that the majority of its attributes belong to the ''occasional" group, reflecting the noisy nature of its high-dimensional generated semantic space. There are hardly any irrelevant or very important attributes. In contrast, AWA2 shows a bimodal tendency but with a clear accumulation of attributes in the “occasional” group as well. This pattern indicates that in this coarse-grained, human-annotated dataset, attributes tend to polarise between very discriminative or almost irrelevant, while still leaving a large middle group with moderate utility. CUB, also human-annotated but fine-grained, presents a strong concentration of attributes in the mid–high frequency range. This suggests that a relatively large set of attributes provides moderate but stable discriminative power, consistent with the subtle inter-class differences that characterise fine-grained datasets. In general, although these three datasets present different distributions, attributes tend to be occasional selected and a little bit polarised in AWA2.

\subsubsection{Fitness and Diversity Convergence Analysis}

\begin{figure*}[htbp]
  \centering
  
  % First row
  \begin{subfigure}[b]{0.45\textwidth}
    \centering
    \includegraphics[width=\linewidth]{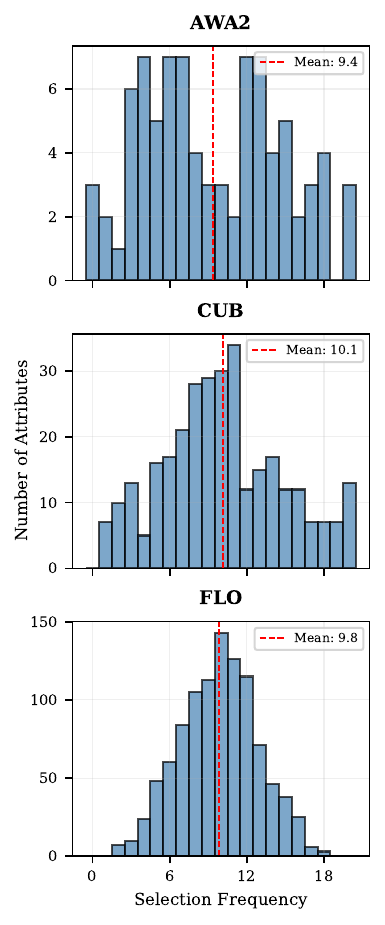}
    \label{fig:att_freq}
    \caption{Attribute Frequencies}
  \end{subfigure}
  \hfill
  \begin{subfigure}[b]{0.45\textwidth}
    \centering
    \includegraphics[width=\linewidth]{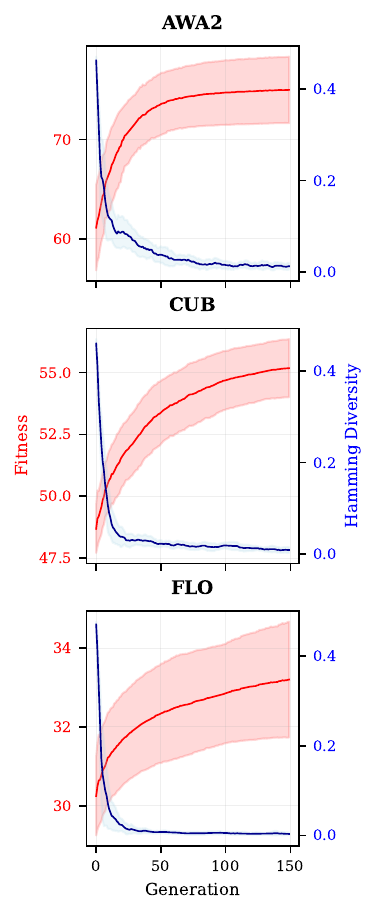}
    \label{fig:fitness-diversity}
    \caption{Diversity-Fitness Curves}
  \end{subfigure} 
  \caption{(a) Attribute selection frequency across $20$ independent runs of the genetic algorithm, showing the stability and relevance of attributes in different datasets. (b) Evolution of fitness and diversity over generations, illustrating the typical explosive, refinement, and plateau phases as well as the progressive loss of diversity.}

  \label{fig:genetic-behaviour}
\end{figure*}

Figure \ref{fig:genetic-behaviour} (b) shows the fitness and diversity convergence curves of the executions through generations. The fitness curve is computed from the validation accuracy obtained via the cross-validation, whereas the diversity convergence curve is computed as the average of the Hamming distance among all pairs of individuals across all generations. 

Regarding the fitness curve, all three datasets exhibit a three-phase behaviour. The first is an explosive phase, in which most of the fitness improvement is achieved. The second is a refinement phase, where fitness still increases but more slowly, and individual runs start to diverge. Finally, in the plateau phase, fitness remains almost flat, showing that the search has reached local optima. In CUB and FLO the refinement phase continues with a clear upward trend, while in AWA2 the curve stabilises much earlier. These differences can be explained by the nature and quality of their semantic spaces. While AWA2 with a small number of attributes saturates early, datasets with more high-dimensional semantic spaces exhibit a longer and more unstable refinement since the algorithm must filter a more noise and redundant information.

Across datasets, solutions show high variance in terms of achieved fitness, but low diversity within each run. In practice, this means that every independent execution converges to a different local optimum and then keeps exploring its neighbourhood. This behaviour reflects the strong sensitivity of the algorithm to stochasticity: while the global pattern (explosion–refinement–plateau) is consistent, the final outcome of each execution can differ widely. In particular, FLO has the largest variability in fitness outcomes, which indicates that the genetic search is particularly unstable in this high-dimensional generated space.

% \textcolor{blue}{Creo que el parrafo este lo puedo quitar porque me centro dataset por dataset y tampoco es que diga nada importante...}
% From these curves, we can also draw conclusions about the datasets themselves. In AWA2 (coarse-grained, human-annotated), the fitness gain saturates early, suggesting that a small number of attributes dominate the discrimination and the search space is relatively simple. In CUB (fine-grained, human-annotated), the slower but steady growth indicates that useful combinations of attributes are harder to identify and improvements require longer exploration. In FLO (fine-grained, generated-based), the strong variability and higher instability reflect the noisy and redundant nature of the learned semantic space. 

\subsubsection{Analysing the impact of Cross-Validation}

\begin{figure}[H]
    \centering
    \includegraphics[width=0.8\linewidth]{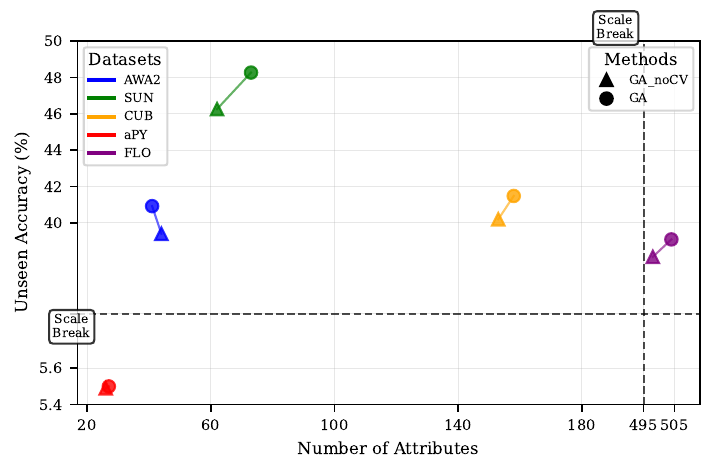}
    \caption{Comparison between GA (with cross-validation, circles) and $GA_{\text{noCV}}$ (single split, triangles) across datasets. Each point shows the number of selected attributes on the $x$-axis and the unseen accuracy ($Acc_{\mathcal{Y}^u}$) on the $y$-axis. Arrows connect paired results within the same dataset. In all cases GA yields higher accuracy with only minor differences in the number of selected attributes.}
    \label{fig:paquita}
\end{figure}

We now assess the impact of the partitioning strategy on the genetic algorithm by comparing GA (with $5$-fold cross-validation) against $GA_{\text{noCV}}$ (single standard predefined split \cite{xian18ugly}). Results in Fig.~\ref{fig:paquita} show that GA consistently achieves higher unseen accuracy than $GA_{\text{noCV}}$ across all datasets, while selecting a very similar number of attributes. The margin is clear in all datasets except for aPY.

This pattern suggests that cross-validation provides a more reliable fitness signal for the evolutionary search: by averaging across folds, it reduces variance and discourages overfitting to a lucky/unlucky split, leading to attribute subsets that generalise better to unseen classes. The attribute counts change only slightly (GA often selects marginally more features, except in AWA2), indicating that CV mainly improves the quality of the selected subsets rather than their size.

\subsection{Explainability}\label{subsec:explainability}

\begin{figure}
    \centering
    \includegraphics[width=0.8\linewidth]{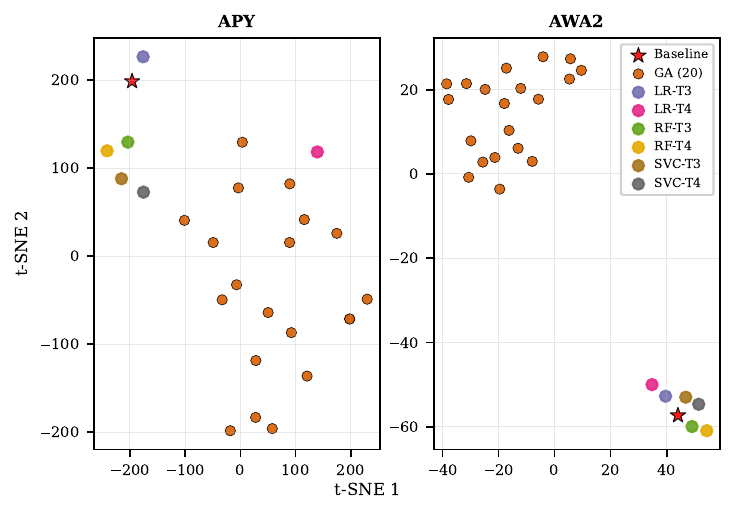}
    \caption{ t-SNE visualisation of solutions using as metric Hamming distance. }
    \label{fig:tsne}
\end{figure}

Regarding explainability, Fig.~\ref{fig:tsne} shows a t-SNE projection of the selected subsets of GA and RFS solutions, where distances are computed using Hamming similarity. For clarity, we only report AWA2 and aPY, since these datasets have the lowest-dimensional semantic spaces and therefore allow a more interpretable visualisation of attribute overlap across methods.

In both datasets, baseline and RFS solutions form compact clusters close to each other, indicating that the subsets selected by RFS remain strongly aligned with the baseline space. This reflects the conservative nature of the ranking-based strategy, which prunes attributes but still preserves most of the original semantic organisation. By contrast, GA solutions are projected further apart, showing that the evolutionary search explores attribute combinations that diverge more substantially from both baseline and RFS. The separation is especially noticeable in AWA2, where GA subsets occupy distant regions of the space, whereas in aPY the overlap is slightly larger but still clearly distinct. 

Overall, these projections confirm the complementary behaviours already discussed. On one hand, RFS yields solutions that are closer to the baseline, refining rather than restructuring the semantic space. On the other hand, GA departs more strongly and proposes alternative combinations that may capture information overlooked in the original design.

%% file: tables/exp_comp.tex
\begin{table}[H]
  \centering
  % Pongo esto porque asumo que aumentará el ancho de la tabla cuando la rellenemos.
  \resizebox{0.8\textwidth}{!}
  {%
    \begin{tabular}{l|cc|cc|cc|cc|cc}
    \hline
     & \multicolumn{2}{c|}{\textbf{AWA2}} 
     & \multicolumn{2}{c|}{\textbf{SUN}} 
     & \multicolumn{2}{c|}{\textbf{CUB}} 
     & \multicolumn{2}{c|}{\textbf{aPY}} 
     & \multicolumn{2}{c}{\textbf{FLO}} \\ 
    \cline{2-11}
     & $Att$ & $Acc_{\mathcal{Y}^{u}}$  
     & $Att$ & $Acc_{\mathcal{Y}^{u}}$  
     & $Att$ & $Acc_{\mathcal{Y}^{u}}$  
     & $Att$ & $Acc_{\mathcal{Y}^{u}}$  
     & $Att$ & $Acc_{\mathcal{Y}^{u}}$  \\
    \hline
    SAE       & 85 & 40.36    
                   & 102 & 47.01    
                   & 312 & 38.79    
                   & 65 & 5.50    
                   & 1024 & 37.66    \\
%    RF-T3          & 75 & 42.53   
%                   & 100 & 46.81 
%                   & 308 & 39.03  
%                   & 47 & \textcolor{red}{\textbf{21.64}}    
%                   & 991 & 37.58  \\
    RFS         & 82 & \textbf{43.98}
                   & 100 & 46.88  
                   & 287 & 39.16   
                   & 43 & \textbf{21.27}
                   & 364 & \textbf{42.6} \\
%    LR-T3          & 70 & 39.72  
%                   & 101 & 47.71
%                   & 293 & 38.96
%                   & 43 & 5.50
%                   & 780 & 38.10    \\
%    RF-T4          & 65 & 42.64   
%                   & 100 & 46.81   
%                   & 293 & 39.50  
%                   & 44 & 21.39
%                   & 873 & 36.97 \\
%    SVC-T4         & 78 & 42.16  
%                   & 98 & 46.60   
%                   & 265 & 39.30      
%                   & 32 & 15.50
%                   & 282 & 42.51 \\
%    LR-T4          & 45 & 39.76
%                   & 95 & 47.15  
%                   & 271 & 39.81   
%                   & 32 & 5.50
%                   & 441 & 36.80   \\
    $GA$          & 41 & 40.92
                   & 73 & \textbf{48.27}
                   & 158 & \textbf{41.48}
                   & 27 & 5.50
                   & 504 & 39.09 \\
%    $GA_{RFS}$         & 38 & 41.59
%                   & 77 & 48.08
%                   & 205 & 41.43
%                   & 24 & 5.50
%                   & - & - \\
    \hline
    \end{tabular}%
  }
    \vspace{0.1cm}
    \caption{Results of RFS, Genetic and Baseline approaches. The number of selected attributes (\textit{Att}) across solutions is shown along with the unseen accuracy ($Acc_{\mathcal{Y}^u}$). The best results obtained in each dataset are highlighted in \textbf{bold}. }
  \label{tab:comparison}
\end{table}